# Person Monitoring by Full Body Tracking in Uniform Crowd Environment

Zhibo Zhang[1], Omar Alremeithi [1], Maryam Almheiri[1], Marwa Albeshr[1], Xiaoxiong Zhang[1], Sajid Javed[1], and Naoufel Werghi[1]

[1] Department of Electrical Engineering and Computer Science, Khalifa University, Abu Dhabi 127788, UAE
`qiuyuezhibo@gmail.com`

**Abstract.** Full body trackers are utilized for surveillance and security purposes, such as person-tracking robots. In the Middle East, uniform crowd environments are the norm which challenges state-of-the-art trackers. Despite tremendous improvements in tracker technology documented in the past literature, these trackers have not been trained using a dataset that captures these environments. In this work, we develop an annotated dataset with one specific target per video in a uniform crowd environment. The dataset was generated in four different scenarios where mainly the target was moving alongside the crowd, sometimes occluding with them, and other times the camera's view of the target is blocked by the crowd for a short period. After the annotations, it was used in evaluating and fine-tuning a state-of-the-art tracker. Our results have shown that the fine-tuned tracker performed better on the evaluation dataset based on two quantitative evaluation metrics, compared to the initial pre-trained tracker.

## 1 Introduction

Object detection is one of the pillars of the Artificial Intelligence and Computer Vision field. In our case, we are interested in person detection and identification. Person detection is where the object matches the semantics of a person and identification is when the person matches an already existing reference template that was obtained prior.

Person detection and identification enable the ability to track a person who is the target. Its application includes surveillance and security such as a person tracking robot [1]. A challenge specific to the area of the Middle East is a uniform environment where men would be wearing white kandora and black abaya for women. For instance, in Saudi Arabia, during the pilgrimage, everyone is wearing white clothes shown in Fig. 1 and a person tracker would not be successful in following a suspicious target due to the high similarity between the crowd. Another example is in the malls and airports in the Middle East where the case of uniform crowd exists and tracking someone in that environment is a difficult task.

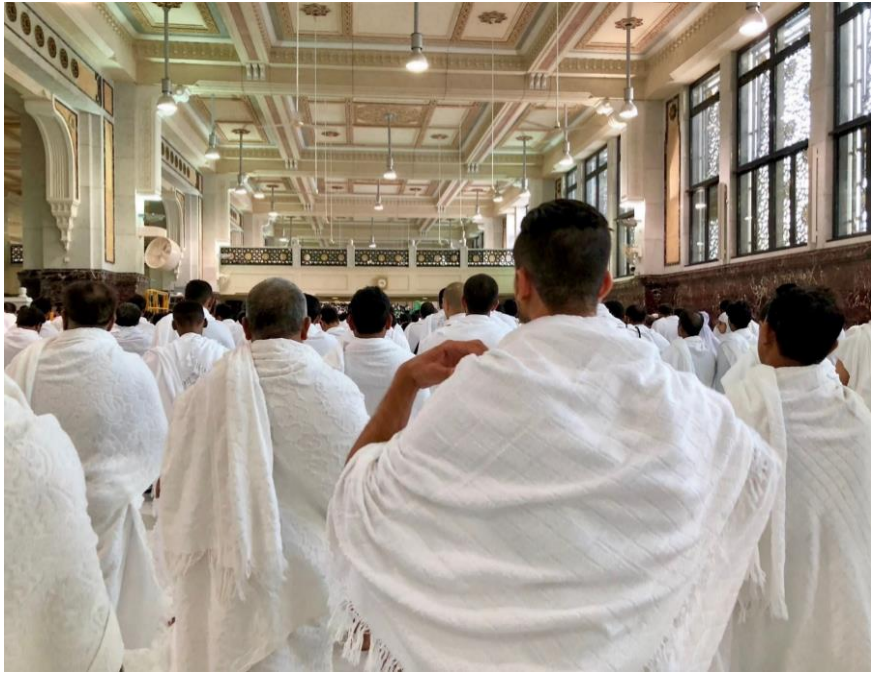

**Fig. 1.** Real-life uniform crowd

Therefore, to address the problem of tracking a person in a uniform crowd environment, we are proposing to generate our own dataset of a uniform crowd environment which we will use to train and deploy a state-of-the-art object tracker. The tracker will be improved and fine-tuned to make it more fitting to the established crowd uniform environment dataset. The results of the fine-tuned tracker based on the established dataset will be compared with the state-of-art algorithms as well. Moreover, the dataset will be publicly available for researchers to use and contribute to [1], with the goal of having a large dataset that is enough to enable person tracking in a uniform crowd environment.

---

[1] https://github.com/qiuyuezhibo/kandora-and-abaya-uniform-tracking-dataset
(we plan to maintain this website as a contribution to the community.)

## 2 Related Work

### 2.1 Siamese trackers

Deep Siamese Networks (SNs) have been widely employed to address generic object tracking as a similarity learning problem [2]. In Visual Object Tracking (VOT), an offline deep network is trained on numerous pairs of target images to learn a matching function, after which an online evaluation of the network as a function takes place during tracking. The Siamese tracker is subdivided into two categories: the template branch, which receives the target image patch from the previous frame as input, and the detection branch, which receives the target image patch from the current frame. Fig. 2 illustrates the tracking pipeline of a standard SN.

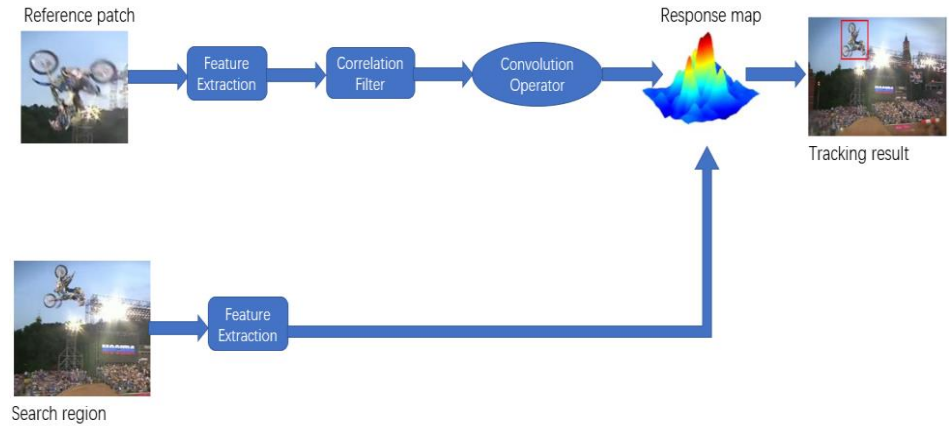

**Fig. 2.** The Siamese tracking pipeline for generic object tracking.

Although Siamese trackers are proven to outperform Discriminative Correlation Filters (DCF)-based trackers in terms of accuracy and efficiency, several limitations to them will be reviewed below along with proposed ways of addressing them.

**Backbone Architectures.** The backbone network is the main element in feature extraction in offline training because of its critical function in obtaining high-level semantic information about the target. Early Siamese trackers were designed based on a modified, fine-tuned AlexNet [3]. However, as modern deeper networks were introduced, SNs were alternatively designed based on such networks (e.g., ResNet [4], VGG-19 [5], and Inception [6]). The results of evaluating VGG-16 and AlexNet models in SINT show a capability difference between the two systems. [7]. Two studies proposed leveraging the powerful ResNet network as a backbone architecture for Siamese trackers which showed excellent performance improvement [8], [9]. Nonetheless, a striking discovery was the absence of performance increase in traditional SNs-based trackers as a result of substituting strong deep architectures directly. [8]. Reasons behind this issue were investigated to be, among others, features padding and receptive fields of neurons, and network strides.

**Offline Training.** Training data plays a dominant role in facilitating the learning process of a powerful matching function in SNs. However, the tracking community has handled this situation well. where immense efforts have been put into compiling large diverse datasets of annotated images and videos [10], [11], [12]. An issue that remains at hand is the inability of the standard Siamese formulation to exploit the appearance of distractor objects during training. To address the matter, hard negative mining techniques were proposed to overcome data imbalance issues by including more semantic negative pairs in the training process. This helped overcome drifting by focusing more on fine-grained representations [13]. Another negative mining technique was the use of an embedding network and the nearest neighbor approximation [11].

**Online Model Update.** In SiamFC [4], the target pattern is established in the first frame and remains constant throughout the video. Failure to update the model means complete reliance of the performance on the SN's matching ability, which poses a huge limitation in scenarios where appearance changes during tracking. Proposed potential methods to solve this issue are Moving Average Update Method [14], [7], [15], Learning Dynamic SN Method [16], Dynamic Memory Network Method [17], Gradient-Guided Method [18], and UpdateNet Method [19].

**Loss Functions.** The loss functions employed within the SNs have an impact on the tracking performance. Whether for regression, classification, or for both tasks, the several types of loss functions can be summarized into the following: Logistic Loss [4], Contrastive Loss [20], Triplet Loss [21], Cross-Entropy Loss [22], Intersection over Union (IoU) Loss [23] and Regularized Linear Regression [24].

**Target State Estimation.** Scale variations are a challenging difficulty that SNs suffer from, where the similarity function does not account for scale changes between images. Multiple Resolution Scale Search Method [25], [26], Deep Anchor-based Bounding Box Regression Method [4], [22], [27], Deep Anchor-free Bounding Box Regression Method [28], [29] are the most popular strategies to address the scaling issue.

### 2.2 Spatio-temporal Transformer Network for Visual Tracking (STARK)

A straightforward benchmark for object detection and tracking is the Encoder-Decoder Transformer which can track objects given spatial information only. The components of the baseline are: Backbone, Encoder-Decoder Transformer, and a Prediction Head for the Bounding Box [30], [31]. An improvement to this baseline is STARK, Spatio-temporal Transformer Network for Visual Tracking, which takes into consideration both the spatial and temporal features. As seen in Fig. 3, STARK has extra components in addition to the components of the baseline; an extra input to the Convolutional Backbone which is the Dynamic Template that provides temporal information, and a Score Head that determines whether the Dynamic Template should be updated or not, and the utilization of different Training and Inference strategies than that of the baseline. It is also important to note that experiments on different benchmarks, using the Spatio-Temporal Tracker, have shown that it is able to perform better than other methods of tracking [29].

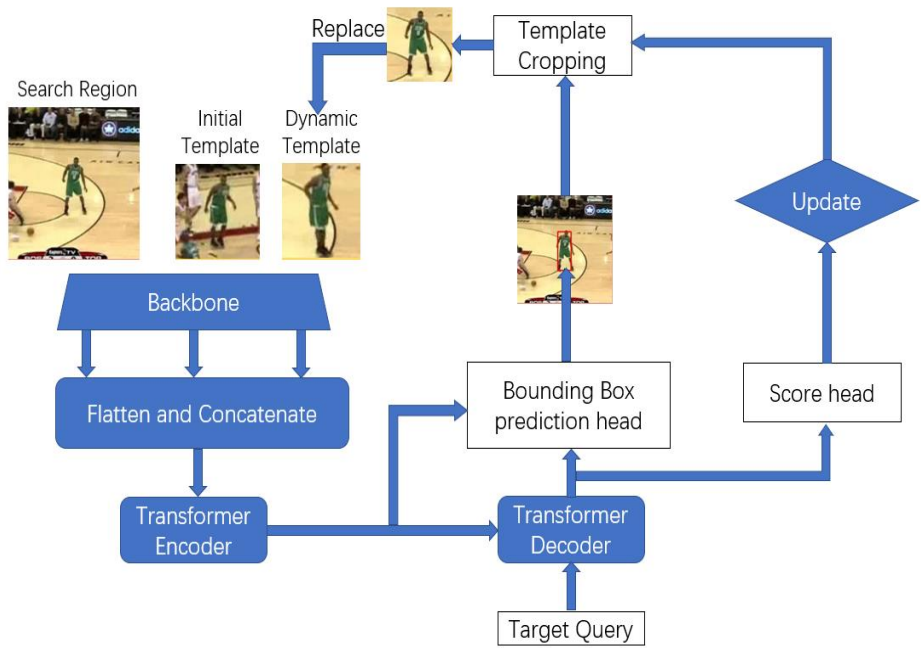

**Fig. 3.** The benchmark STARK pipeline for object tracking.

The training process in STARK is split into two stages: localization, and classification. The split leads to a better solution compared to joint learning. The first stage trains the whole network with the exclusion of the scoring head, while the second stage trains only the score head to learn classification and avoid losing its localization expertise [29].

## 3 Methodology

In this section, in order to satisfy the objective of detecting and tracking a person in a uniform crowd, we will discuss our approaches in the following steps. Section 3.1 presents the process of uniform crowd data collection. Section 3.2 shows how the generated data are annotated with the ground truth using Computer Vision Annotation Tool (CVAT). Section 3.3 describes the splitting of the dataset into training datasets, testing datasets, and validation datasets. Section 3.4 explains the training processes of both the original STARK model and the fine-tuned model.

### 3.1 Collection of Data

The problem of person tracking among a uniform crowd was approached by tackling the issue of not having a ready-made dataset. Therefore, multiple videos were recorded. and each video had a target among a crowd that is characterized by having uniform clothing. In our case, we had four targets: two men, and two women.

In each video, the target does several things: moving around, interacting with other people in the crowd, trying to stay hidden for a while (i.e., vanishing from the camera's view), and trying to be semi-visible.

Similarly, the people in the crowd were also moving around, interacting among themselves or with the target, and trying to block the target from the camera's view.

Another factor that was taken into consideration while recording some of the videos is the lighting. In real life, crowds may be in different lighting settings. To take this into account, the room was initially well lit, and the lighting gradually decreased until the room became dimly lit. Then, the lighting gradually increased to go back to its original status.

Considering as many complex scenarios as possible enables better handling by the tracker of challenging real-life situations, therefore, four particular scenarios were selected to be captured when generating the videos dataset. A description of the scenarios is given below:
1. Target walking among a compact crowd of two/three crisscrossing distractors.
2. Target walking among a relatively dispersed crowd of two/three crisscrossing distractors.
3. Target walking alone, two/three distractors join the target all walking in a straight line, distractors then orderly split away into different paths.
4. Target walking alone, two/three distractors join the target all walking in a straight line, distractors criss-cross with the target then scatter away into different paths.

Each of the four scenarios was recorded twice with every target, once with 2 distractors and once with 3 distractors. Fig. 4 demonstrates selected frames from the recorded videos representing each scenario, where the target is labeled in red.

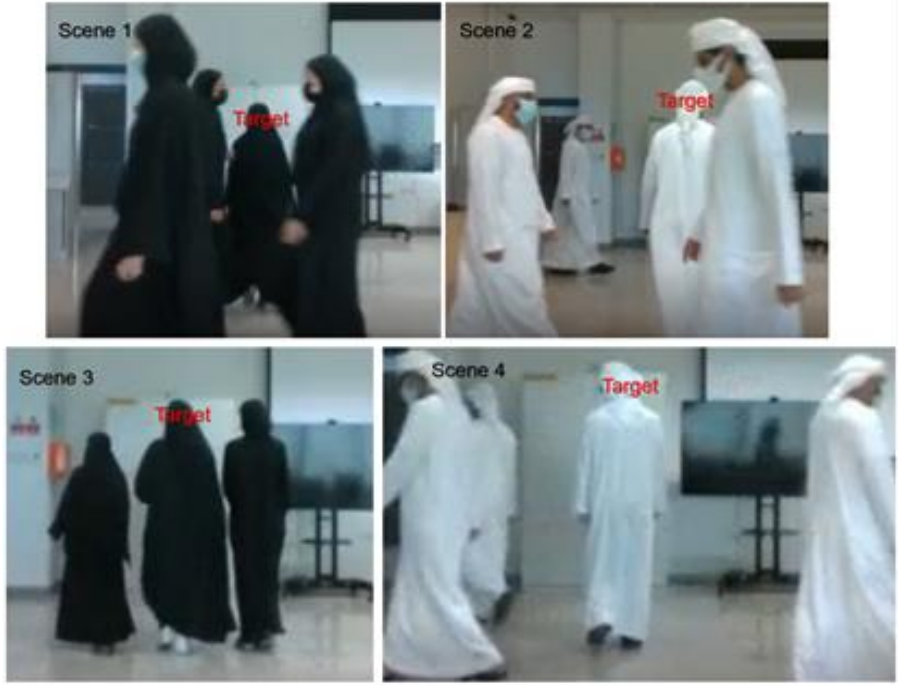

**Fig. 4.** Selected frames from recorded scenarios.

A total of 41 videos were recorded, 32 of them of the four scenarios mentioned above. A summary is shown in Table 1. Targets 1 and 2 represent the men. Whereas Targets 3 and 4 represent the women.

**Table 1.** Summary of Recorded Videos.

| Target | Number of videos | Number of people in crowd |
|---|---|---|
| 1 | 10 | 2, 3 |
| 2 | 10 | 2, 3 |
| 3 | 12 | 1, 2, 3, 4 |
| 4 | 9 | 2, 3, 4 |

### 3.2 Annotation of generated data

To create a dataset dedicated to training and performance evaluation, twenty videos were annotated: five videos for each Target. The annotation was done through the Computer Vision Annotation Tool (CVAT).

The annotations were done by encapsulating the target in a rectangular box and checking the corresponding attributes. The attributes depend on the status of the Target as seen in the frame. Since the Targets and crowd are constantly moving, both the size of the rectangle and the attributes must be updated in each frame.

Two attributes were chosen for the project: Occlusion and Out of View. The first attribute is checked whenever the Target is obscured by a member of the crowd but can still be seen. Whereas the second attribute is checked whenever the Target is fully unseen from the camera's point of view.

Fig. 5 shows examples of annotated frames with each possible combination of attributes (Occlusion and Out of View) for the target.

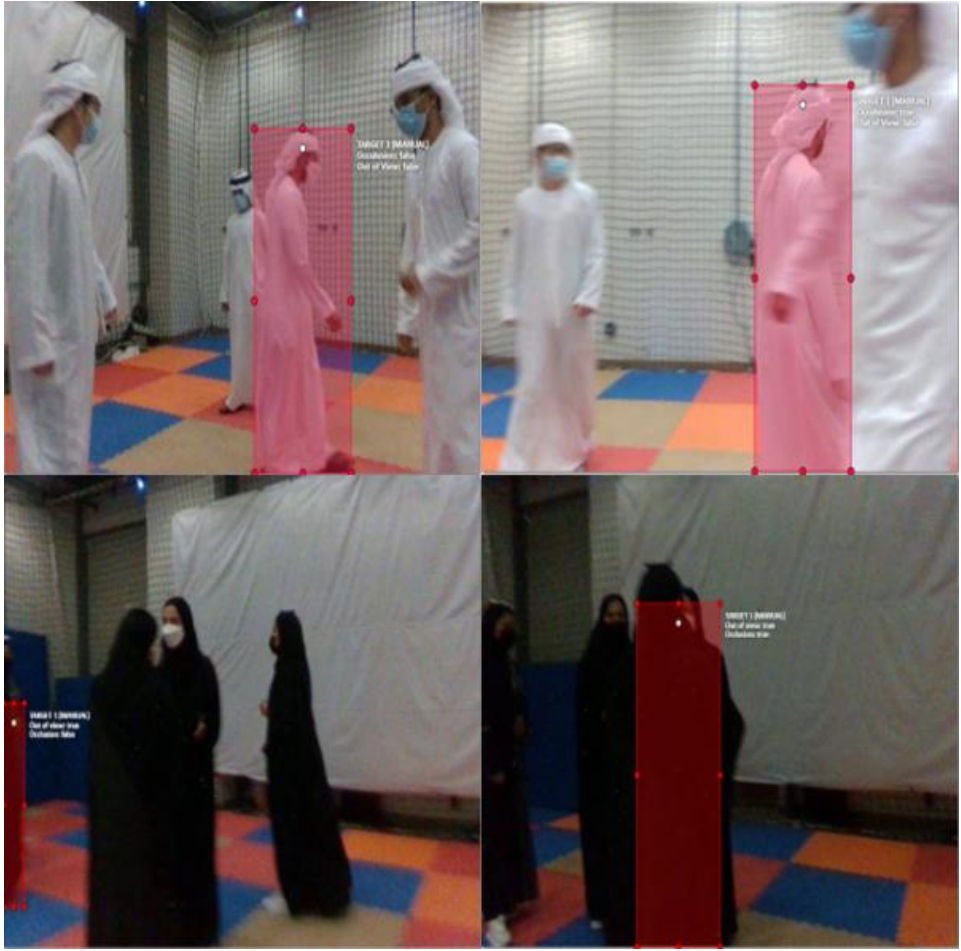

**Fig. 5.** Examples of every attribute combination

The following table contains the number of annotated frames for each Target.

Table 2. Annotated Frames Number details.

| Target | Number of annotated frames |
|---|---|
| 1 | 2358 |
| 2 | 2381 |
| 3 | 2161 |
| 4 | 2481 |
| Total annotated frames | 9381 |

It is worth mentioning that additional videos of different targets that were recorded by other groups were utilized to expand the final dataset to ensure better performance and reduce the risk of data leakage when evaluating the performance. Table 3 shows a summary of said videos where Targets 5 and 6 are in a female crowd, and Target 7 is in a male crowd.

Table 3. Summary of Additional Videos.

| Target | No. of videos | No. of people in the crowd | No. of annotated frames |
|---|---|---|---|
| 5 | 8 | 2, 3 | 769 |
| 6 | 8 | 2, 3 | 706 |
| 7 | 4 | 2, 3 | 775 |
| Total annotated frames | | | 2250 |

### 3.3 Splitting of Dataset

To go about with the training and testing, the generated dataset was split into training, validation, and testing sets. Table 4 summarizes the videos allocated to each set. The testing process was performed 3 separate times in which different targets were involved at a time; once with Dataset 1 (Targets 1, 2, 3 & 4), once with Dataset 2 (Targets 5 & 6), and once with Dataset 3 (Target 7).

Table 4. Splitting of generated Dataset.

| | Number of videos | Targets involved | Number of frames |
|---|---|---|---|
| Training | 12 | 1, 2, 3, 4 | 8051 |
| Validation | 2 | 3, 4 | 497 |
| Testing 1 | 8 | 1, 2, 3, 4 | 1330 |
| Testing 2 | 8 | 5, 6 | 1475 |
| Testing 3 | 4 | 7 | 775 |

### 3.4 Training process

We use STARK-ST50 as the basic model and test the impact of various components on various datasets. Res-Net-50 serves as the backbone for the baseline tracker STARK-ST50.

The backbone is pre-trained using ImageNet. During training, the Batch Norm layers are frozen. Backbone qualities from the fourth stage are combined with a stride of 16. The transformer architecture is the same as DETR, with six encoder levels and six decoder layers, including multi-head attention layers (MHA) and feed forwarding (FFN). The MHA has 8 heads with 256 widths, but the FFN has hidden units with 2048 widths. The dropout ratio is set to 0.1. A light FCN composed of five stacked Conv-BN-ReLU lay-ers serves as the bounding box prediction head. The classification head is a three-layer perceptron with 256 hidden units in each layer.

The minimal training data unit for STARK-ST is one triplet, consisting of two templates and one search image. The whole training process of STARK-ST consists of two stages, which take 500 epochs for localization and 50 epochs for classification, respectively. Each epoch uses $6 \times 10^4$ triplets. The network is optimized using AdamW optimizer and weight decay $10^{-4}$. The loss weights $\lambda_{L1}$ and $\lambda_{iou}$ are set to 5 and 2 respectively. Each GPU hosts 16 triplets, hence the mini-batch size is 128 triplets. The initial learning rates of the backbone and the rest parts are $10^{-5}$ and $10^{-4}$ respectively. The learning rate drops by a factor of 10 after 400 epochs in the first stage and after 40 epochs in the second stage.

We compare the original model with the fine-tuned model based on different datasets. The former is the STARK-ST50 model that is open-sourced at [32] which has been trained for 500 epochs in stage 1 and 50 epochs in stage 2, whereas the Fine-tuned model uses pre-training: we took a pre-trained model with 500 epochs, and we trained for it for 10 epochs using stage 1. Afterward, the resulting model was trained for 5 epochs using stage 2.

## 4 Results and Discussion

As mentioned in Section 3, the fine-tuned model implemented pre-trained techniques to make the model perform better, including 10 epochs using stage 1 and 5 epochs using stage 2. We selected the third epoch from stage 2 based on the validation loss and training loss shown in Fig. 6 where the third epoch had the minimum validation loss.

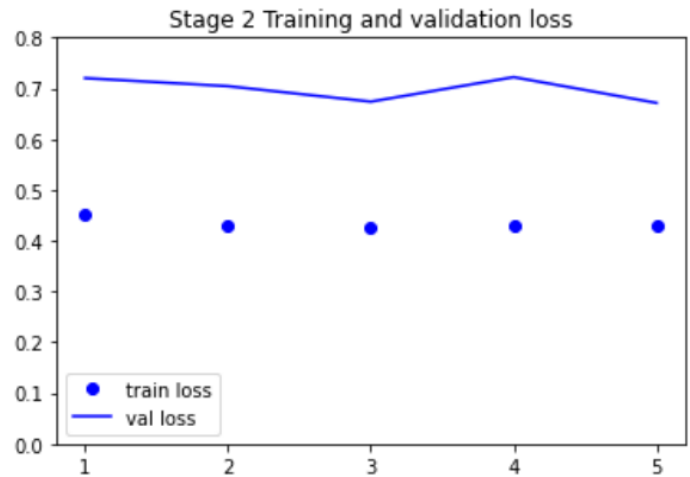

**Fig. 6.** Training and validation loss in stage 2

The visualization results of the comparison are shown in Fig. 7. The red rectangular demonstrates the ground truth of the target person in that video annotated in CVAT, the blue rectangular indicates the tracking results of the original model, and the green rectangular shows the tracking results of the fine-tuned model. From the frames shown in Fig. 7 and the generated videos, it is obvious that the fine-tuned model performs much better than the original model. However, to show the conclusion quantitatively, we introduce the concepts of Precision and AUC(Area Under the ROC Curve).

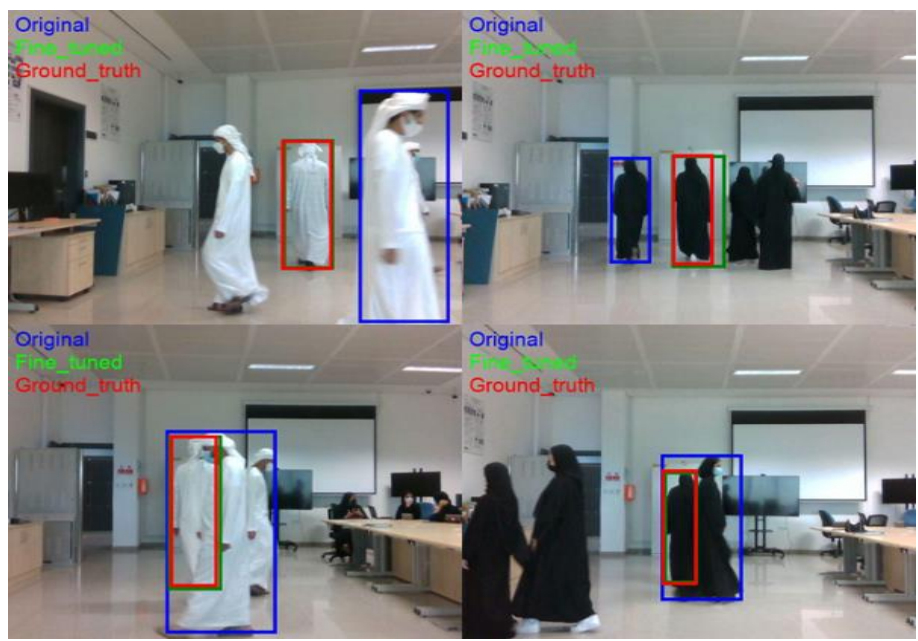

**Fig. 7.** Visualization of Datasets 1, 2, and 3 respectively, including men and women.

Precision is the distance between measure values, or how many decimal places exist at the end of a specific measurement. The definition of Precision is defined as the following equation:

$$Precision = \frac{True\ Positive}{True\ Positive + False\ Positive} \quad (1)$$

A receiver operating characteristic curve (ROC curve) is a chart that illustrates a classification model's performance across all categorization levels. This graph shows two parameters: the True Positive Rate and the False Positive Rate. The term AUC is an acronym for "Area Under the ROC Curve." In other words, AUC is the whole two-dimensional area beneath the entire ROC curve from (0,0) to (1,1). (1,1).

From Fig. 7, it is obvious that the results of the Fine-tuned model are superior to the results of the Original state-of-art STARK model in most figures. However, we deployed the standard performance metrics AUC (Area under the ROC Curve), and Precision. A ROC curve (receiver operating characteristic curve) is a graph that depicts a classification model's performance across all categorization levels. The True Positive Rate and False Positive Rate are plotted on this graph. Table 5 displays the actual performance of the various models on various datasets.

**Table 5.** Performance of STARK-ST50 with and without pre-training on datasets.

| Model | Dataset | AUC | Precision |
|---|---|---|---|
| STARK-ST50 (Original) | Dataset 1 | 49.51% | 50.22% |
| STARK-ST50 (Fine-tuned) | Dataset 1 | 82.24% | 88.62% |
| STARK-ST50 (Original) | Dataset 2 | 58.28% | 56.47% |
| STARK-ST50 (Fine-tuned) | Dataset 2 | 76.21% | 82.56% |
| STARK-ST50 (Original) | Dataset 3 | 56.47% | 58.63% |
| STARK-ST50 (Fine-tuned) | Dataset 3 | 81.00% | 87.56% |

It is noticeable that for every dataset, the performance of our proposed fine-tuned model is superior to the performance of the original state-of-art STARK-ST50 model, which demonstrates the significance of the two-stage pre-training process. It is also noteworthy that the performance between the datasets is similar. Although Dataset 2 and 3 were strictly consisting of either males wearing white kandoora or females wearing black abaya, respectively, the results show that our tracking methods would not be influenced by people wearing different colors of uniforms. Therefore, after a fair comparative analysis between the proposed fine-tuned model and the state-of-art STARK algorithm in the literature, the performance of the proposed fine-tuned model is superior to the state-of-art STARK algorithm in terms of AUC and Precision on different sets of the developed sets of uniform crowd datasets.

## 5  Conclusion

Through our work, we have developed a dataset that encapsulates uniform crowd environments in four different scenarios. The dataset was annotated manually by bounding the target with a box and setting the appropriate attributes. A pre-trained STARK-ST50 model was fine-tuned on the dataset. Afterward, the original STARK-ST50 model and the fine-tuned model were evaluated on the dataset using the AUC and Precision metrics. Furthermore, two additional datasets were provided by colleagues for more evaluation. Results have shown that the fine-tuned model performed significantly better than the original model since the fine-tuned model is experienced with the environment. We will also investigate the related body segmentation challenge given the constraints in such datasets [33,34].

However, there are still some limitations in this work. Firstly, the proposed datasets include targets among 3 or 4 people, so this may pose a problem regarding scalability. Thus, more complicated scenarios with larger uniform crowds should be developed. Besides, although the proposed fined-tuned model outperformed the start-of-art algorithms in terms of crowd monitoring, the tracking results should be further improved for better monitoring performance. Therefore, we propose two parallel steps for future work. Firstly, the development of more datasets incorporating more complex scenarios, including the monitoring of large crowded events, with more attributes can be used for analyzing the results and providing deep insight into the model's behavior. Secondly, a modification of the architecture of STARK should be made to make it more resilient and aware of uniform crowds.